\documentclass[11pt]{article}

\usepackage{acl}

\usepackage{times}
\usepackage{latexsym}

\usepackage[T1]{fontenc}

\usepackage[utf8]{inputenc}

\usepackage{microtype}

\usepackage{inconsolata}

\usepackage{cite}
\usepackage{xspace}
\usepackage{multirow}
\usepackage{booktabs}
\usepackage{xcolor}
\usepackage{caption}
\usepackage{graphicx}
\usepackage{amsmath}
\usepackage[frozencache,cachedir=minted-cache]{minted}
\usepackage{soul}
\usepackage{amssymb}
\usepackage{amsmath,amsfonts,bm}

\def\eqref#1{equation~\ref{#1}}

\def\1{\bm{1}}

\def\va{{\bm{a}}}

\def\vs{{\bm{s}}}

\DeclareMathAlphabet{\mathsfit}{\encodingdefault}{\sfdefault}{m}{sl}
\SetMathAlphabet{\mathsfit}{bold}{\encodingdefault}{\sfdefault}{bx}{n}

\newcommand{\ie}{\textit{i.e.}}
\newcommand{\eg}{\textit{e.g.}}
\newcommand{\sr}{Self-Rubrics\xspace}
\newcommand{\cf}{Contrastive Filtering\xspace}
\newcommand{\llama}{LLaMA\xspace}
\newcommand{\palm}{PaLM\xspace}
\definecolor{ao}{rgb}{0.0, 0.5, 0.0}
\definecolor{cadmiumgreen}{rgb}{0.0, 0.42, 0.24}
\usepackage{pifont}
\newcommand{\xmark}{\ding{55}}

\newcommand{\method}{CodecLM\xspace}
\newcommand{\methodabbr}{CodecLM\xspace}

\title{CodecLM: Aligning Language Models with Tailored Synthetic Data}

\author{Zifeng Wang$^\dagger$, Chun-Liang Li$^\dagger$, Vincent Perot$^*$, Long T. Le$^\dagger$, \\ \bf Jin Miao$^\ddagger$, Zizhao Zhang$^\ddagger$, Chen-Yu Lee$^\dagger$, Tomas Pfister$^\dagger$ \\
$^\dagger$Google Cloud AI Research, $^\ddagger$Google Cloud AI, $^*$Google Research\\
\texttt{\{zifengw, chunliang, vperot, longtle,} \\ 
\texttt{jinmiao, zizhaoz, chenyulee, tpfister\}@google.com}
}

\begin{document}
\maketitle
\begin{abstract}
Instruction tuning has emerged as the key in aligning large language models (LLMs) with specific task instructions, thereby mitigating the discrepancy between the next-token prediction objective and users' actual goals. To reduce the labor and time cost to collect or annotate data by humans, researchers start to explore the use of LLMs to generate instruction-aligned synthetic data. 
Recent works focus on generating diverse instructions and applying LLM to increase instruction complexity, often neglecting downstream use cases. It remains unclear how to \emph{tailor} high-quality data to elicit better instruction-following abilities in different target instruction distributions and LLMs. To this end, we introduce \textbf{\method}, a general framework for adaptively generating high-quality synthetic data for LLM alignment with different downstream instruction distributions and LLMs. Drawing on the Encode-Decode principles, we use LLMs as codecs to guide the data generation process. 
We first \emph{encode} seed instructions into metadata, which are concise keywords generated on-the-fly to capture the target instruction distribution, and then \emph{decode} metadata to create tailored instructions. We also introduce \sr and \cf during decoding to tailor data-efficient samples. Extensive experiments on four open-domain instruction following benchmarks validate the effectiveness of \methodabbr over the current state-of-the-arts.

\end{abstract}

\section{Introduction}

\begin{figure}[t]
    \centering
    \includegraphics[width=0.95\linewidth]{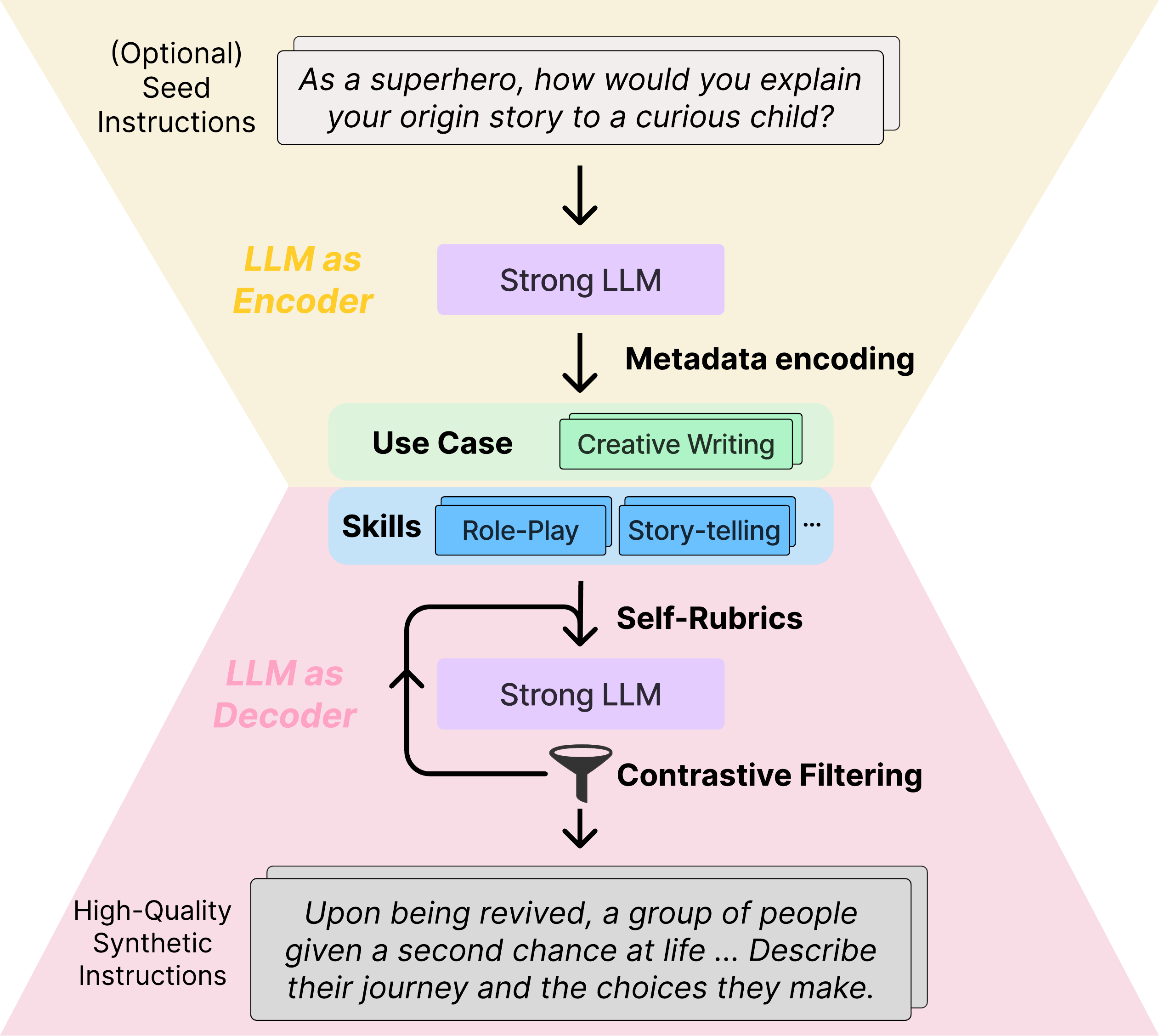}
    \vspace{-2mm}
    \caption{Overview of \method. We first \textbf{encode} seed instructions into metadata to capture the underlying distribution of instructions. This metadata is then \textbf{decoded} through \sr and \cf to tailor high-quality synthetic instructions that are aligned with the target instruction distribution. Intermediate instructions and responses are omitted in the figure for clarity.}
    \label{fig:overview}
    \vspace{-5mm}
\end{figure}

Large language models (LLMs) have exhibited remarkable capabilities across a wide array of natural language processing (NLP) tasks \citep{brown2020language,ouyang2022training,OpenAI2023GPT4TR,anil2023palm}. In particular, LLMs can be trained for improved instruction-following through various methods, including fine-tuning on human-annotated data~\citep{touvron2023llama,bai2022training} or extracted knowledge from stronger LLMs~\citep{wang2022self,alpaca,vicuna2023,peng2023instruction}. Recent progress in this area highlights the critical role of high-quality data in enhancing LLMs' instruction-following capabilities~\citep{zhou2023lima,kopf2023openassistant,chen2023alpagasus}. However, acquiring such data through human annotation remains cost-prohibitive and difficult to scale, hindering further progress.

As an alternative solution to human annotation, recent work explores generating instruction-response pairs for LLM alignment by prompting them with example data or prompts and iteratively refining the results~\citep{honovich2022unnatural,wang2022self,li2023self,xu2023wizardlm}. %
While these methods are effective at generating diverse and complex instructions for LLM alignment broadly, real-world applications often prioritize tailoring the LLM to specific downstream tasks such as individual enterprise applications or personal assistant agents~\citep{OpenAI2023IntroducingGPTs}, which often involve different instruction distributions. This desideratum for task-specific alignment brings us to a core question for data synthesis: \emph{how can we tailor synthetic data to align LLMs for different instruction-following tasks?}

Specifically, current data synthesis approaches fall short of providing effective solutions for task-specific LLM alignment. While prior works by
\citet{wang2022self} and \citet{xu2023wizardlm} emphasize diversity and complexity as hallmarks of high-quality data, these approaches stumble when facing different downstream tasks that may involve specific instruction distributions. A diverse dataset for one task might not effectively cover the instruction distribution for another. Furthermore, the definition of ``complex'' instructions can be subjective and vary across tasks. To complicate matters further, an LLM might excel at some seemingly complex instructions while struggling with others that appear simple according to human-crafted criteria. These limitations underscore the need for a unified data synthesis framework that can generate tailored data to align LLMs on specific downstream tasks.

In this work, we present a novel framework, \textbf{\method}, which systematically generates tailored high-quality data to align LLMs for different downstream tasks. A high-level overview of \methodabbr is shown in Figure~\ref{fig:overview}. Inspired by the principles of Encode-Decode process~\citep{kramer1991nonlinear,kingma2013auto}, we leverage a strong LLM as a codec to ``encode'' seed instructions from our target task into instruction \emph{metadata} and then ``decode'' the metadata into tailored synthetic instructions.
The metadata serves as a word-level abstraction of the input instruction distribution, including the \emph{use case} and \emph{skills} for effective instruction following. It can be automatically generated by encoding seed instructions, or directly provided by users with a high-level anticipation of the downstream task. 

Once the metadata is extracted, we then  ``decode'' them to generate tailored instructions. We begin by prompting a LLM with the metadata as constraints, creating basic instructions. To elevate the instruction quality, we introduce \emph{\sr}. It samples appropriate actions from strong LLMs to make the basic instruction more complex or challenging based on the rubrics it generates for different metadata.
Intuitively, a general knowledge QA instruction about math would differ in complexity rubrics from one in creative writing about sports. With self-generated rubrics and actions based on metadata, the strong LLM crafts instructions that better align the target LLM with specific knowledge required for the downstream task. We can run \sr iteratively to control the instruction complexity, similar to ~\citet{xu2023wizardlm}, and finally generate the corresponding responses. 

We also introduce \emph{\cf} during decoding to further identify the most effective instruction-response pairs by leveraging the quality discrepancy between the target and a stronger LLM. This strategy identifies two key instruction sets: (a) those the target LLM struggles with, pushing it to improve in its weak areas for more significant gains, and (b) those the target LLM excels at, feeding them back into the \sr process for improved data efficiency.
\cf serves as a response-level analogy of contrastive decoding~\citep{li2022contrastive}.

\method sets a new state-of-the-art on four open-domain instruction-following benchmarks with various LLM choices, demonstrating its effectiveness in LLM alignment for diverse instruction distributions.

\section{Related Work}
\textbf{Instruction Tuning for LLM Alignment.} Tuning LLM to faithfully follow instructions and align with diverse human preferences remains a significant challenge~\citep{efrat2020turking}. Early research primarily focused on cross-task generalization, where models were fine-tuned on various public NLP datasets to improve performance on diverse tasks~\citep{raffel2020exploring,wei2021finetuned,aribandi2021ext5,victor2022multitask,chung2022scaling}.
More recently, researchers have extended instruction tuning to open-domains, characterized by a wider range of formats and task types. This shift has been driven by crowdsourcing human-generated instruction-response pairs~\citep{ouyang2022training,kopf2023openassistant,zhou2023lima} and LLM-generated data~\citep{alpaca,vicuna2023}.
Unlike prior work, \method presents a unique approach for tailoring synthetic data to specific downstream tasks without human annotation, utilizing the concept of instruction metadata.

\noindent \textbf{Data Generation for Instruction Tuning.} To address the high cost of human annotation for high-quality instruction-response pairs, several studies advocate for automating the data generation process~\citep{schick2021generating,liu2022wanli,meng2023tuning}. Leveraging the in-context learning~\citep{brown2020language} ability of LLMs, ~\citet{wang2022self,honovich2022unnatural} prompt LLMs with seed instructions to generate synthetic ones. These are then fed to stronger LLMs, \eg, ChatGPT, to generate responses for training the target (often smaller) LLM~\citep{alpaca}. As a representative work, WizardLM~\citep{xu2023wizardlm}, designs a fixed set of human-crafted operations to increase complexity of instructions and control difficulty of generated data. \citet{zhao2023preliminary,zhou2023lima} further confirm the importance of instruction complexity for LLM alignment through empirical studies. Different from these works that rely on pre-defined rules without considering the downstream tasks, \method enables automatically tailoring instructions for different downstream tasks and target LLMs. We also introduce \sr and \cf to further identify the most effective instruction-response pairs.

\noindent \textbf{Distillation.} Alternatively, tuning the target LLM with responses generated from another LLM can be viewed as knowledge distillation~\citep{hinton2015distilling,beyer2022knowledge}. However, our focus remains on instruction generation, while still being flexible to readily integrate with existing distillation techniques~\citep{hsieh2023distilling,liang2023less}.

Finally, we discuss some of the most relevant recent work. AttrPrompt~\citep{yu2023large} leverages LLM as attributed data generator by extracting attributes within instructions. However, it focuses solely on classification tasks and requires human intervention for attribute selection. In contrast, our work focuses on the broader context of aligning LLMs to follow open-domain instructions, eliminating the need for human efforts. MSP~\citep{chen2023mixture} utilizes trainable soft prompts to control generation, but requires gradient access to the LLM. Our method, on the other hand, is readily compatible with black-box LLMs that only offer API access for high-quality data generation. SteerLM~\citep{dong2023steerlm} analyzes quality-related aspects of responses, instead of the instructions, to capture human preference. Therefore, SteerLM can be used alongside \methodabbr as a parallel approach for enhancing response quality.

\begin{figure*}
    \centering
    \includegraphics[width=0.99\linewidth]{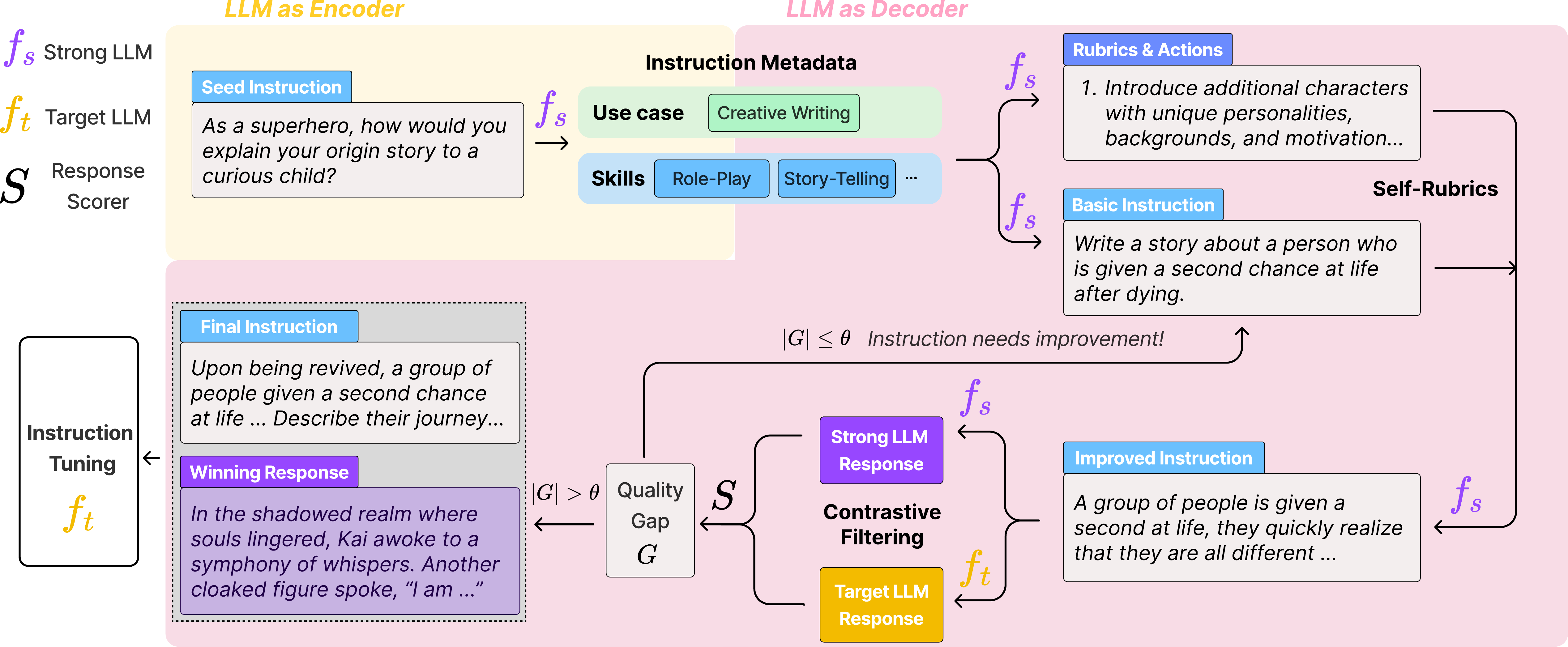}
    \vspace{-3mm}
    \caption{Overview of the proposed \methodabbr. First, the strong LLM $f_s$ encodes the seed instruction into instruction metadata, specifying its use case and skills required for responses. Next, $f_s$ decodes metadata into basic instructions. Meanwhile, \sr leverages $f_s$ to generate rubrics and actions to improve the basic instruction, tailoring them for the downstream task. Finally, Contrastive Filtering uses a scoring function $S$ to compares $f_s$ and $f_t$'s responses. The most effective pairs are selected for aligning the LLM, while less effective instructions are sent for further improvement. In this figure, the strong LLM's response is winning against the target one's, so we select the corresponding pair for instruction tuning the target LLM.}
    \label{fig:detailed_overview}
    \vspace{-3mm}
\end{figure*}

\section{Problem Statement}
We study the open-domain instruction following problem~\citep{wang2022self,alpaca,xu2023wizardlm}, where instructions vary in input format and tasks. Specifically, we consider two practical scenarios: (1) Starting with a given set of $n$ seed instructions $\mathcal{D}_s = \{I_i\}_{i=1}^{n}$, each drawn from some underlying distribution $P_I$. For our experiments, we create a set of seed instructions using a held-out validation set. Practically, such instructions can be collected from the usage traffic of users. (2) In the absence of seed instructions, but with prior knowledge of downstream tasks, we directly start with a given set of instruction metadata $\mathcal{M}$ (see Section~\ref{sec:encoder} for definition). The latter scenario is especially useful for end users who lack existing instruction data but wish to jumpstart LLM tailored to specific applications, similar to the concept of GPTs~\citep{OpenAI2023IntroducingGPTs}. 

We focus on the first scenario for clarity, though the second can be derived similarly by leveraging an LLM as the encoder (Section~\ref{sec:encoder}). Our goal is to generate a set of high-quality instruction-response pairs $\mathcal{D}_g = \{(I^{'}_{j}, R^{'}_{j})\}_{j=1}^{m}$, using a strong LLM $f_s$, and then use $\mathcal{D}_g$ to fine-tune the target LLM $f_t$. We evaluate the performance of the fine-tuned LLM $f_t$ on test instructions from the target distribution $P_I$, to which we are aligning.

\section{\method}
We propose \method, a general framework for generating high-quality instruction-response pairs tailored to different downstream tasks and LLMs, eliminating the need for human annotation. See Figure~\ref{fig:detailed_overview} for method overview. %

\subsection{LLM as Codec for Instructions} \label{sec:encoder}
In this section, we introduce the concept of using a strong LLM as a codec, \ie, both encoder and decoder, for instruction generation.

\textbf{LLM as Encoder with Instruction Metadata.}
We begin by encoding the given seed instructions $\mathcal{D}_s = \{I_i\}_{i=1}^{n}$ into instruction \emph{metadata} $\mathcal{M}$, \ie, keywords that capture the underlying target instruction distribution. Inspired by the task pool by~\citet{wang2022self} and the post-hoc analysis on skill distribution by~\citet{xu2023wizardlm}, we define the metadata as encompassing two key aspects: \emph{use case} and \emph{skills}. Use case describes the intended task (\eg, question answering or creative writing), while Skills are the knowledge the LLM required to have to successfully respond to the given instruction (\eg, algorithms or communication). Skills are often generalizable to different use cases. Therefore, each instruction has a single use case and may involve multiple skills. 
To extract this metadata, we leverage the strong LLM $f_s$ following the prompt template in Figure~\ref{fig:metadata}, Appendix~\ref{app:prompts}. While richer definitions are possible based on finer-grained instruction-following metrics~\citep{zhou2023instruction}, we prioritize use case and skills for their broad applicability across diverse instruction distributions. Future work can explore extending this metadata further.

For each instruction $I_i$, we extract the corresponding use case $u_i$ and set of skills $\vs_i$. We then have the set of metadata as $\mathcal{M} = \{(u_i, \vs_i)\}_{i=1}^{n}$. Instructions may share or partially overlap in their $u_i$'s and $\vs_i$, reflecting the distribution of tasks and capabilities within the seed instructions. 
Use cases and skills are generated on-the-fly, not limited to some predefined sets, enabling broader applicability. However, we can always provide such constraints with our prior knowledge, or even directly write out metadata without any seed instructions.

\textbf{LLM as Decoder for Instruction Generation.} Given the metadata $\mathcal{M}$, we decode metadata into synthetic instructions, following a generation and tailoring paradigm. For each use case and skills pair in $\mathcal{M}$, we list them as constraints to prompt the strong LLM $f_s$ to generate multiple instructions. Therefore, the generated instructions are for the given use case, and require the given skills to be responded. 
Moreover, to prevent the LLM from generating repetitive instructions, we encourage its generation to be diverse in the prompt, and do not provide any demonstrations that the LLM might copy from. The example prompt template for generating basic instructions is in Figure~\ref{fig:base_generation}, Appendix~\ref{app:prompts}.  Continuing the decoding process, we then tailor the basic instructions for more effective alignment through \sr (Section~\ref{sec:self_rubrics}) and \cf (Section~\ref{sec:contrastive_filtering}). %

\subsection{Instruction Tailoring via Self-Rubrics} \label{sec:self_rubrics}
Metadata-conditioned instructions lay the groundwork for aligning the target LLM to desired tasks. Studies suggest that more complex instructions can improve alignment performance~\citep{xu2023wizardlm,zhao2023preliminary}. A common practice is to involve human experts crafting general guidance to complicate instructions, such as adding reasoning steps or constraints. However, this one-size-fits-all strategy falls short for diverse instructions. Tailoring guidance to different tasks, like solving calculus problems versus writing news articles, requires distinct approaches.

Therefore, we introduce Self-Rubrics, which leverages the strong LLM to tailor instructions by adjusting their complexity according to the extracted metadata.
Self-Rubrics first guides the LLM to generate metadata-specific rubrics for assessing instruction complexity. Then, informed by these rubrics, the LLM generates a corresponding set of actions to enhance the instruction's complexity.
For metadata $(u_i, \vs_i)$, the corresponding set of generated actions is $\va_i$. Our generated actions are more domain-specific, and unambiguous than generic rules crafted by human, making the complicated instructions better tailored towards the target distribution captured by the metadata. For example, for the use case of ``business plan development'' and skills of ``market research and planning'', generic rules like ``add reasoning steps'' is vague and inappropriate. On the contrary, \sr is able to generate actions like ``add SWOT analyisis'' and ``include comparison with market competitors'' (see Appendix~\ref{app:case_study} for the full details) to complicate the instruction. 
The prompt template to generate rubrics and actions for instruction improvement is shown in Figure~\ref{fig:rubric_action_generation}, Appendix~\ref{app:prompts}.

With the obtained actions $\{\va_i\}_{i=1}^{n}$, we can iteratively prompt $f_s$ to complicate the basic instructions, following the prompt template in Figure~\ref{fig:instruction_improvement}. %
We randomly sample an action $\va_i$ from the multiple actions generated for a pair of use case and skills. This design choice not only enables controlled complexity~\citep{xu2023wizardlm}, but also prevents potential confusion between different actions for the LLM. %

\subsection{Instruction Selection via Contrastive Filtering} \label{sec:contrastive_filtering}
While Self-Rubrics tailors complex instructions based on instruction metadata, not all instructions are equally effective for instruction tuning, regardless of their complexity~\citep{chen2023alpagasus,zhou2023lima}. %
Intuitively, exposing the target LLM to instructions it finds challenging can effectively identify its areas for improvement. Therefore, it is crucial to select the most impactful instructions for aligning the target LLM. 

We therefore introduce \cf, a method to select the instructions that can effectively enhance the target LLM $f_t$. For clarity, we define the space of all natural language sequences as $\mathcal{N}$. We have the strong LLM $f_s: \mathcal{N} \to \mathcal{N}$, the target LLM $f_t: \mathcal{N} \to \mathcal{N}$, and a scoring function $S: \mathcal{N} \to \mathbb{R}$ to evaluate response quality. 
In practice, $S$ is obtained by reusing the strong LLM $f_s$ with a prompt template (Figure~\ref{fig:cf_scorer}, Appendix~\ref{app:prompts}) adapted from the Vicuna pairwise evaluation template~\citep{alpaca,vicuna2023}.
To mitigate potential position bias, we average the scores obtained by exchanging the positions of two responses~\citep{vicuna2023}. 
We observe using $f_s$ for scoring works quite well in practice, so we prioritize this option for simplicity.
Given an input instruction $I \in \mathcal{N}$, we obtain responses from both LLMs as $f_s(I)$ and $f_t(I)$, respectively. We then define the \emph{quality gap} $G: \mathcal{N} \to \mathbb{R}$ between these responses to estimate the \emph{effectiveness} of the instruction: $G(I) = S(f_s(I)) - S(f_t(I))$.

The quality gap metric $G$ reflects how much the target LLM benefits from the strong LLM for each instruction $I$.
As demonstrated in Figure~\ref{fig:detailed_overview}, here are two possible cases: (1) $|G(I)| > \theta$, where $\theta \in \mathbb{R}$ is a certain threshold. This indicates that: Either the strong LLM has a much better response than the target LLM, we add $(I, f_s(I))$ to our high-quality instruction-response pool $\mathcal{D}_g$ to fill the gap; Or rarely, the target LLM gives much better response than the strong LLM, we add $(I, f_t(I))$ to $\mathcal{D}_g$ as as an implicit regularization to keep the target LLM's desirable behavior to certain instructions. (2) $|G(I)| \leq \theta$, where the quality of responses from both LLMs is similar, so learning from $I$ does not lead to much gain. We then send $I$ to the next Self-Rubrics iteration for further improvement.

\cf complements \sr to select effective instruction-response pairs by calibrating the target LLM's instruction-following capability with the strong LLM's. Analogous to Constrastive Decoding~\citep{li2022contrastive} at response-level, \cf can also be regarded as LLM-feedback~\citep{madaan2023self} with the interaction of two LLMs. While we adopt the strong LLM as scoring function to measure the quality gap, our framework can be compatible with and potentially benefit from the advances in more reliable and comprehensive scoring and feedback systems~\citep{lee2023applying}, and we leave it as promising future work. 

\section{Experiments}
We conduct comprehensive experiments to evaluate \methodabbr using different LLMs on multiple representative benchmarks, closely following well-established evaluation settings for open-domain instruction following in prior work~\citep{xu2023wizardlm,chen2023alpagasus}. 
We also conduct a case study in Appendix~\ref{app:case_study} to illustrate how \method tailors an instruction step by step.

\subsection{Evaluation Benchmarks}
We evaluate \method on four widely-used open-domain instruction-following benchmarks with diverse instruction distributions to reduce evaluation bias. Our test benchmarks include Evol-Instruct~\citep{xu2023wizardlm}, Vicuna~\citep{vicuna2023}, Self-Instruct~\citep{wang2022self} and Koala~\citep{geng2023koala}.
To complement the evaluation, we also evaluate on two standard NLP benchmarks MMLU~\citep{hendrycks2020measuring} and BBH~\citep{suzgun2022challenging} in Appendix~\ref{app:additional_benchmarks}.
Please refer to Appendix~\ref{app:benchmark} for benchmark details.

\subsection{Baseline Methods}
We compare our method against state-of-the-art data generation approaches for instruction tuning. For fair comparison, we provide all methods the same LLM backbones when possible. Moreover, we control the number of instruction-response pairs the same for all methods to ablate the effect of data quantity.
Baseline methods include \textbf{Self-Instruct}~\citep{wang2022self}, \textbf{Alpagasus}~\citep{chen2023alpagasus}, \noindent~\textbf{Tree-Instruct}, \textbf{WizardLM}~\citep{xu2023wizardlm}, and \textbf{WizardLM+}, an enhanced version of WizardLM using the same basic instructions generated from \method as seed instructions. Baseline details are presented in Appendix~\ref{app:baseline_details}.

\subsection{Experiment and Evaluation Details}
\textbf{LLM Backbones.} We adopt \llama-based~\citep{touvron2023llama} and \palm-based~\citep{anil2023palm} LLMs as our target LLMs in our experiments. For \llama-based target LLMs, we use Gemini-Pro~\citep{team2023gemini} as the strong LLM, and \llama-7B, -13B as the target LLMs. For \palm-based target LLMs, we use {text-unicorn} as the strong LLM, and {text-bison} as the target LLM. \palm-based models and Gemini-Pro are accessible through Google Cloud API\footnote{\url{https://cloud.google.com/vertex-ai}}.

\noindent \textbf{Implementation Details of \methodabbr.} We split all benchmarks into 20\% validation set and 80\% evaluation set. We extract the instruction metadata from the validation set, see Appendix~\ref{app:implementation_details} for more details. Depending on the specified total data size, we prompt the strong LLM to generate equal number of base instruction per metadata. We generate 500-8000 synthetic data throughout the experiments. We generate 4 rubrics and corresponding actions. At each iteration, we randomly choose 1 action for improving instruction. We run \sr at most 4 iterations. For \cf, We set the scoring scale to 10 and the filtering threshold to 3 for all experiments. We align these configurations with~\citet{xu2023wizardlm}
and leave more detailed rationales of these configurations, additional hyperparameter settings, and training details in Appendix~\ref{app:implementation_details}-\ref{app:training_details}.

\begin{table*}[htbp]
\small
\centering
\caption{Results with \llama-based target models on four open-domain instruction following benchmarks. Each method trains a target model based on \llama-7B or -13B, and compares against the strong model, Gemini-Pro. The reported metric Capacity Recovery Ratio (\%), $\texttt{CRR} = \frac{\texttt{wins} + \texttt{ties}}{\texttt{total comparisons}}$. Larger CRR means better performance.
}
\vspace{-2mm}
\scalebox{0.88}{
\begin{tabular}{l||cccc||cccc}
\toprule
\multirow{2}[4]{*}{\textbf{Methods}} & \multicolumn{4}{c||}{\textbf{\llama-7B vs. Gemini-Pro}} & \multicolumn{4}{c}{\textbf{\llama-13B vs. Gemini-Pro}} \\
\cmidrule{2-9}      & \textbf{Evol-Ins.} & \textbf{Vicuna} & \textbf{Koala} & \textbf{Self-Ins.} & \textbf{Evol-Ins.} & \textbf{Vicuna} & \textbf{Koala} & \textbf{Self-Ins.} \\
\midrule
Self-Instruct & 72.02 & 81.25 & 67.78 & 65.87 & 75.69 & 86.25 & 77.22 & 69.05 \\
Alpagasus & 75.23 \scriptsize{\textcolor{cadmiumgreen}{(+3.2)}} & 81.25 \scriptsize{\textcolor{cadmiumgreen}{(+0.0)}} & 71.11 \scriptsize{\textcolor{cadmiumgreen}{(+3.3)}} & 70.24 \scriptsize{\textcolor{cadmiumgreen}{(+4.4)}} & 79.82 \scriptsize{\textcolor{cadmiumgreen}{(+4.1)}} & 87.50 \scriptsize{\textcolor{cadmiumgreen}{(+1.3)}} & 77.78 \scriptsize{\textcolor{cadmiumgreen}{(+0.6)}} & 71.03 \scriptsize{\textcolor{cadmiumgreen}{(+2.0)}} \\
Tree-Instruct & 75.23 \scriptsize{\textcolor{cadmiumgreen}{(+3.2)}} & 81.25 \scriptsize{\textcolor{cadmiumgreen}{(+0.0)}} & 72.78 \scriptsize{\textcolor{cadmiumgreen}{(+5.0)}} & 68.65 \scriptsize{\textcolor{cadmiumgreen}{(+2.8)}} & 82.57 \scriptsize{\textcolor{cadmiumgreen}{(+6.9)}} & 87.50 \scriptsize{\textcolor{cadmiumgreen}{(+1.3)}} & 80.56 \scriptsize{\textcolor{cadmiumgreen}{(+3.3)}} & 79.37 \scriptsize{\textcolor{cadmiumgreen}{(+10.3)}} \\
WizardLM & 74.31 \scriptsize{\textcolor{cadmiumgreen}{(+2.3)}} & 76.25 \scriptsize{\textcolor{red}{(-5.0)}} & 65.56 \scriptsize{\textcolor{red}{(-2.2)}} & 71.43 \scriptsize{\textcolor{cadmiumgreen}{(+5.6)}} & 82.11 \scriptsize{\textcolor{cadmiumgreen}{(+6.4)}} & 86.25 \scriptsize{\textcolor{cadmiumgreen}{(+0.0)}} & 78.89 \scriptsize{\textcolor{cadmiumgreen}{(+1.7)}} & 76.19 \scriptsize{\textcolor{cadmiumgreen}{(+7.1)}} \\
WizardLM+ & 75.69 \scriptsize{\textcolor{cadmiumgreen}{(+3.7)}} & 83.75 \scriptsize{\textcolor{cadmiumgreen}{(+2.5)}} & 68.33 \scriptsize{\textcolor{cadmiumgreen}{(+0.6)}}  & 72.22 \scriptsize{\textcolor{cadmiumgreen}{(+6.4)}} & 84.40 \scriptsize{\textcolor{cadmiumgreen}{(+8.7)}} & 88.75 \scriptsize{\textcolor{cadmiumgreen}{(+2.5)}} & 81.11 \scriptsize{\textcolor{cadmiumgreen}{(+3.9)}} & 79.76 \scriptsize{\textcolor{cadmiumgreen}{(+10.7)}} \\
\methodabbr (ours) & \bf 79.82 \scriptsize{\textcolor{cadmiumgreen}{(+7.8)}} & \bf 88.75 \scriptsize{\textcolor{cadmiumgreen}{(+7.5)}} & \bf 74.44 \scriptsize{\textcolor{cadmiumgreen}{(+6.7)}} & \bf 78.17 \scriptsize{\textcolor{cadmiumgreen}{(+12.3)}} & \bf 86.70 \scriptsize{\textcolor{cadmiumgreen}{(+11.0)}} & \bf 90.00 \scriptsize{\textcolor{cadmiumgreen}{(+3.8)}} & \bf 82.22 \scriptsize{\textcolor{cadmiumgreen}{(+5.0)}} & \bf 83.33 \scriptsize{\textcolor{cadmiumgreen}{(+14.3)}} \\
\bottomrule
\end{tabular}%
}
\label{tab:results-llama}%
\end{table*}%

\noindent \textbf{Evaluation.} 
Assessing how well LLMs follow instructions is complex, arising from the fact that an instruction has various valid responses, and the challenge of replicating human evaluation. Recent advances in automatic evaluation on instruction following~\citep{dubois2023alpacafarm,zheng2023judging} demonstrate that LLM-based evaluators are scalable, explainable, and consistent with human evaluations. Therefore, we adopt widely-used Vicuna pairwise evaluator~\citep{vicuna2023} based on ChatGPT to compare the response quality from two LLMs for its accessibility in price and efficiency. The evaluation prompt template is in Figure~\ref{fig:vicuna_judge}, Appendix~\ref{app:prompts}. We include GPT-4 based evaluation results in Appendix~\ref{app:gpt4} to demonstrate the consistency of LLM-based evaluators. To mitigate position bias that the LLM evaluator may have, we conduct every evaluation twice by exchanging response orders. A response is considered better only if it wins twice. Following~\citep{chen2023alpagasus}, we set the temperature to 0.0 to reduce evaluation randomness, and left other parameters as default.

Similar to prior work~\citep{xu2023wizardlm,zhao2023preliminary}, we compute the total ratio of wins and ties of a target LLM against the strong LLM, to indicate how much model capacity the target LLM recovers from the strong LLM (often treated as the upper bound performer). CRR simplifies the combinatorial pairwise comparisons between all target LLMs. We name the metric as \emph{Capacity Recovery Ratio} (CRR), where $\texttt{CRR} = \frac{\texttt{wins} + \texttt{ties}}{\texttt{total comparisons}}$. In experiments, we observe that the number of ties often dominates the number of wins, since the strong LLM is much capable than the target model. So we do not put additional weights on wins in the calculation. To demonstrate CRR faithfully reflects model performance, we show the exact number of wins, ties and losses in Appendix~\ref{app:detailed_comparison} on Evol-Instruct. We would like to emphasize our focus on the gap in CRR between different methods instead of the absolute value, since the absolute value may based on the specific LLM evaluator we choose. %

\subsection{Open-Domain Instruction Following} \label{sec:main_results}

\textbf{Results with \llama-based Target LLMs.} Table~\ref{tab:results-llama} summarizes the performance of \methodabbr and the comparing baselines with 2000 synthetic data for instruction tuning. All methods are trained on \llama-7B or -13B as the target LLM and compared against Gemini-Pro, the strong LLM that generates the data. \methodabbr outperforms comparing methods consistently on all benchmarks, with two target LLMs of different sizes. The consistently superior performance of \method highlights its generalizability to different downstream instruction distributions and target LLMs. Both Tree-Instruct and variants of WizardLM focus on the importance of instruction complexity, however, their performances are not always better than Alpagasus with simple instructions, especially with larger target LLM. This observation indicates that the effectiveness of data cannot be solely determined by instruction complexity, and validates the motivation of our design of \sr and \cf. Moreover, the win of WizardLM+ over WizardLM confirms the efficacy of instruction distribution matching via instruction metadata. When shifting the target LLM from \llama-7B to -13B, all methods get a significant performance boost, which accords with prior discoveries on scaling model size~\citep{wei2021finetuned}.

\noindent \textbf{Results with \palm-based Models.} Table~\ref{tab:results-palm} summarizes the results of \methodabbr and the best performing baselines in \llama-based experiments. We generate 1000 synthetic data due to computation budget. Since text-bison is a proprietary model that has been aligned with various techniques including instruction tuning, we also include it as a baseline approach. Interestingly, text-bison obtains strong performance across different benchmarks. Both Alpagasus and WizardLM+ underperform text-bison, suggesting it is non-trivial to improve upon a well-tuned LLM continually. \methodabbr, on the contrary, outperforms text-bison in most cases, thanks to our core designs that adaptively tailor high quality data pairs to improve the target LLM.

\begin{table}[t!]
\small
\setlength{\tabcolsep}{3pt} %
\centering
\caption{CRR Results on \palm-based models. Each method trains a target model based on text-bison, and compares against the strong model, text-unicorn. 
}
\vspace{-3mm}
\scalebox{0.86}{
\begin{tabular}{l||cccc}
\toprule
\multirow{2}[4]{*}{\textbf{Methods}} & \multicolumn{4}{c}{\textbf{text-bison vs. text-unicorn}} \\
\cmidrule{2-5}      & \textbf{Evol-Ins.} & \textbf{Vicuna} & \textbf{Self-Ins.} & \textbf{Koala} \\
\midrule
text-bison & 87.16 & 81.25 & \bf 74.21 & 77.47  \\
Alpagasus & 82.11\scriptsize{\textcolor{red}{(-5.1)}} & 81.25 \scriptsize{\textcolor{cadmiumgreen}{(+0.0)}}  & 67.86 \scriptsize{\textcolor{red}{(-6.4)}} & 73.33 \scriptsize{\textcolor{red}{(-4.1)}}  \\
WizardLM+ & 84.40 \scriptsize{\textcolor{red}{(-2.8)}}  & 78.75 \scriptsize{\textcolor{red}{(-2.5)}}  & 69.44 \scriptsize{\textcolor{red}{(-4.8)}} & 73.89 \scriptsize{\textcolor{red}{(-3.6)}} \\
\methodabbr (ours) & \bf 88.53 \scriptsize{\textcolor{cadmiumgreen}{(+1.4)}} & \bf 86.25 \scriptsize{\textcolor{cadmiumgreen}{(+5.0)}} & 72.22 \scriptsize{\textcolor{red}{(-2.0)}} & \bf 80.56 \scriptsize{\textcolor{cadmiumgreen}{(+3.1)}} \\
\bottomrule
\end{tabular}%
}
\label{tab:results-palm}%
\vspace{-3mm}
\end{table}%

\subsection{Ablation Study} \label{sec:ablation}
\begin{table}[t]
    \small
    \centering
    \caption{Ablation study of \methodabbr's core designs. All components contribute to the final performance.}
    \vspace{-3mm}
    \scalebox{0.85}{
        \begin{tabular}{ccc||c}
        \toprule
        \bf Metadata &  \bf \sr & \bf\cf & \bf CRR \\
        \midrule
        \xmark & \xmark & \xmark & 72.02 \\
        \checkmark & \xmark & \xmark & 75.23 \\
        \checkmark & \checkmark & \xmark & 77.52 \\
        \checkmark & \checkmark & \checkmark & 79.82 \\
        \bottomrule
        \end{tabular}
    }
    \label{tab:ablation}
    \vspace{-3mm}
\end{table}
In this section, we conduct comprehensive ablation studies to empirically explore the effectiveness of \methodabbr. We mainly conduct experiments with \llama-7B model as the target LLM, Gemini-Pro as the strong LLM, and report the CRR on the Evol-Instruct benchmark.

\noindent \textbf{Effectiveness of Core Designs.} 
We show component-wise contributions in our framework in Table~\ref{tab:ablation}. The 1st row has the result from Self-Instruct as a baseline; In the 2nd row, we only align the LLM with basic instructions from instruction metadata; We gradually add \sr and \cf in the 3rd and 4th rows, respectively. %
We clearly observe that every component contributes to the final performance. Interesting, the performance of using basic instructions from metadata is even on par with that of WizardLM+ in Table~\ref{tab:results-llama}. This observation indicates that human-crafted strategies for complicating instructions may not fit different types of instructions. On the contrary, \sr adaptively generates instruction improving actions based on different metadata, resulting in better tailored instructions for the target LLM. Further improvements from \cf demonstrate that selected data are indeed more effective for alignment. 

\begin{figure}
    \centering
    \includegraphics[width=0.9\linewidth]{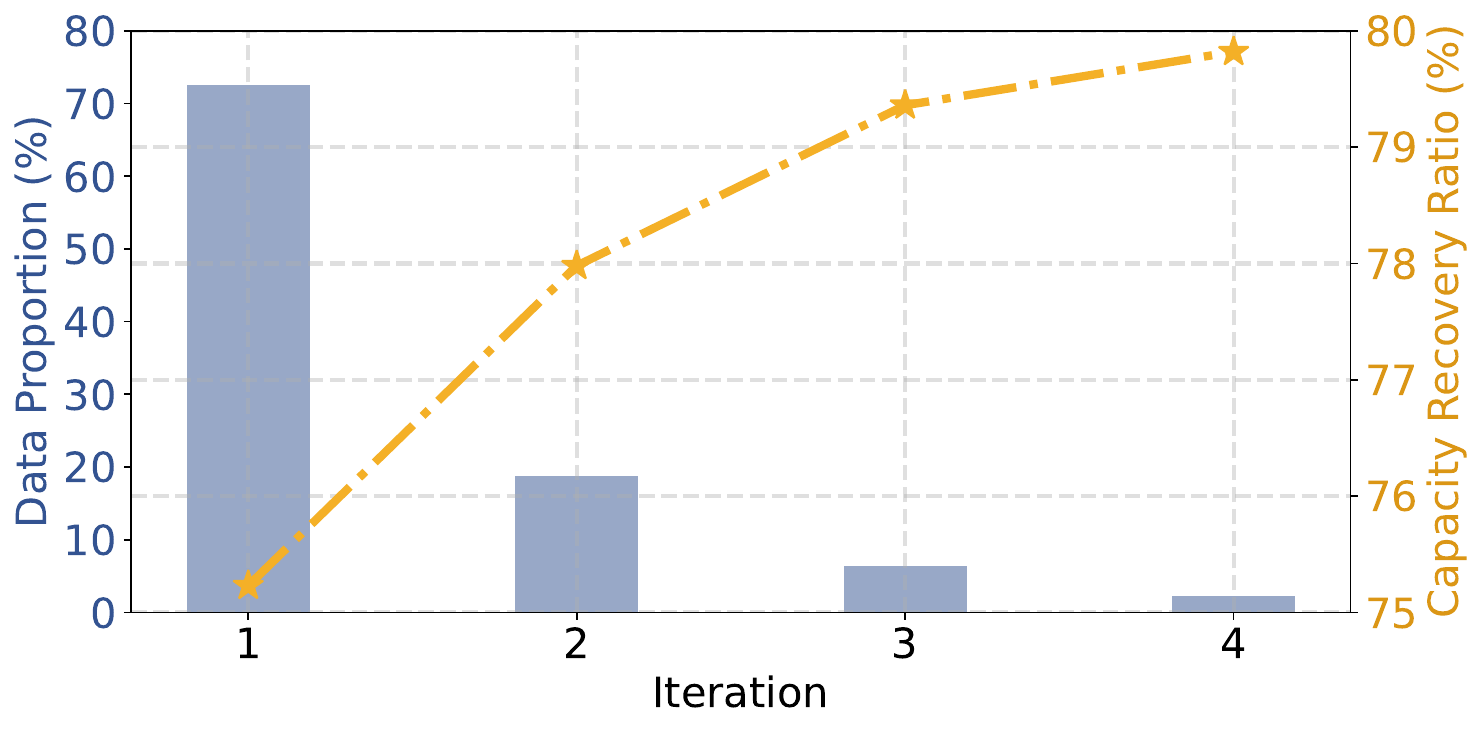}
    \vspace{-3mm}
    \caption{Data proportion from each iteration and the corresponding CRR performance at each iteration.} %
    \label{fig:iteration}
    \vspace{-3mm}
\end{figure}

\noindent \textbf{Effect of Number of Iterations.} We demonstrate the effect of number of \method iterations in Figure~\ref{fig:iteration}. In particular, we count the proportion of data from each iteration in all synthesized data $\mathcal{D}_g$ and show it in the blue bar chart with left y-axis. We also draw the target model performance in CRR after training on the synthetic data up until the current iteration in the yellow line chart with right y-axis. From the data proportion bar chart, we observe that more than $70\%$ of the data comes from the first iteration. This indicates \cf successfully collects less complex yet challenging instructions, which are critical for building up the instruction-following ability of the target LLM. Starting from the second iteration, the data proportion gets increasingly small. However, similar to the \emph{less is more for alignment} observation~\citep{zhou2023lima}, high-quality and more complex instructions indeed contribute to the final performance despite less in quantity.

\begin{figure}
    \centering
    \includegraphics[width=0.9\linewidth]{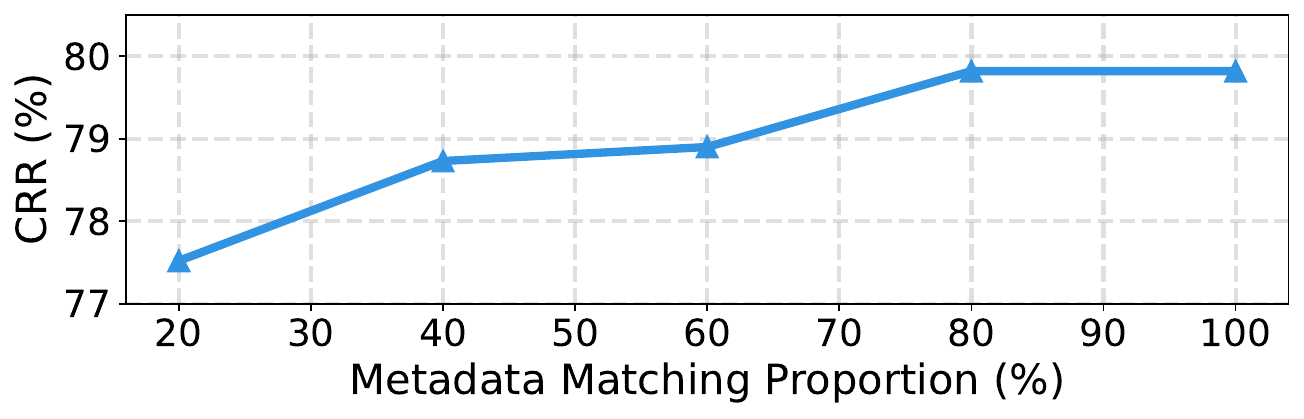}
    \caption{Metadata matching proportion vs. CRR.} %
    \label{fig:metadata_threshold}
    \vspace{-5mm}
\end{figure}

\noindent \textbf{Exploration on Distribution Matching.} As shown by previous results, generating metadata extracted from the downstream instruction distribution indeed helps. However, in practice, the extracted or human-written metadata may not be able to precisely characterize the instruction distribution. Therefore, it is necessary to explore the performance of \methodabbr when the distribution represented by instruction metadata does not fully match the test distribution. As the true test distribution is complicated and not known as a prior, we approximate various extent of distribution matching by random subsampling from the set of metadata $\mathcal{M}$. 
To control the effect of data quantity, we keep the total number of instruction-response pairs the same for each case. For example, when subsampling $20\%$ of $\mathcal{M}$, we prompt the strong LLM to generate 5 times more instructions for each metadata accordingly. The result is shown in the upper part of Figure~\ref{fig:metadata_threshold}, and we did observe the trend that the better instruction metadata captures the underlying distribution, the better performance the target LLM can achieve. Moreover, when the metadata matching proportion is equal or greater than $60\%$, we obtain close performance as the fully-matched result. This observation highlights \methodabbr's robustness under potential instruction metadata mismatch.

\begin{figure}
    \centering
    \includegraphics[width=0.9\linewidth]{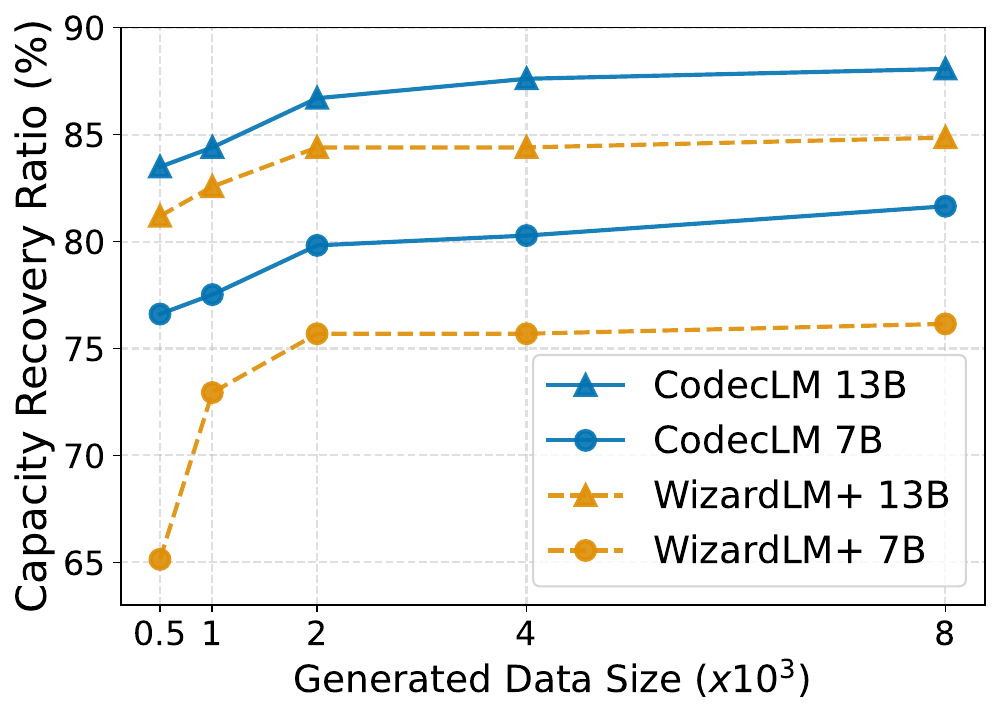}
    \vspace{-3mm}
    \caption{Scaling with model size and data quantity.}
    \label{fig:scaling}
    \vspace{-1mm}
\end{figure}
\textbf{Scaling with Model Size and Data Quantity.} To explore how our method scales with different synthetic data quantities and model sizes, we conduct experiments by comparing \methodabbr with WizardLM+, the most competitive baseline. The experiment results on Evol-Instruct with \llama-7B and -13B as the target LLM are presented in Figure~\ref{fig:scaling}. Both methods get increasingly better performance with more synthetic data and larger target models. \methodabbr consistently outperforms WizardLM+ under all cases, demonstrating its great data efficiency and scalability. We expect the gain will gradually diminish after we generate more than 8k synthetic data, due to the intrinsic ability gap between the target models and the strong LLM.

\section{Conclusion}
In this work, we propose \method to tailor synthetic data for LLM alignment with different target instruction distributions and LLMs. We show that \method effectively captures the underlying instruction distribution via instruction metadata, and further tailor the most effective instruction-response pairs through \sr and \cf. \method provides a potent solution towards adapting LLMs for customized uses, without the necessity of human annotation. We believe \method serves as a general framework for targeted LLM alignment, which opens the door to multiple promising research directions within the framework, such as richer metadata definition, better prompt design, and more reliable LLM-based scorer. \method can also benefit from orthogonal research fields, and we continue the discussion in Ethical Considerations and Limitations sections.

\clearpage

\section*{Ethical Considerations}
Although \methodabbr serves as an effective data synthesis framework for LLM alignment, we should also reflect on the ethical impact of our work. Our method leverages LLMs to generate instruction-response pairs. Similar to human annotators who might make unconscious mistakes during the data annotation process, LLMs also sometimes generate unethical, toxic or misleading instructions and responses~\citep{bender2021dangers}. Moreover, as we train a target LLM using the generated data, the resulting instruction-tuned LLM might also carry the bias and fairness issues~\citep{gallegos2023bias} from the original model. Although we conducted manual inspection as specified in Appendix~\ref{app:implementation_details}, in practice, we should adopt existing techniques ~\citep{Detoxify,thakur2023language} to detoxify and mitigate bias from LLMs used in \methodabbr, and design more strict inspection and filtering rules to clean up the generated data. Due to the flexibility of our framework, we envision future progress in the domain of reducing bias and fairness issues can be complementary to \methodabbr.

\section*{Limitations}
We acknowledge the limitations of \methodabbr from the following aspects to inspire future research opportunities in the field of LLM alignment.

First of all, as discussed in the Ethical Considerations, our method requires a strong LLM to generate the data, so the performance of our method depends on the quality of the LLM and may inherit bias and fairness issues from it. On the other hand, \method can benefit from stronger LLMs improved with advanced bias-reducing and fairness-enhancing approaches.

Secondly, as an orthogonal direction, our method did not explore robustness of the instruction-tuned model towards adversarial attacks such as prompt injection~\citep{liu2023prompt} and jailbreaking~\citep{zou2023universal}. In practice, we should apply adversarial defense techniques~\citep{jain2023baseline} accordingly to the instruction-tuned LLM from our method.

Moreover, we mainly use LLM-based automatic evaluation methods following recent works in data synthesis for alignment. Although recent studies~\citep{vicuna2023,dubois2023alpacafarm} demonstrate LLM-based evaluation is largely consistent with human evaluation, the scalability and reliability of LLM-based evaluators still have room for improvements. Although we include some standard benchmark results in Appendix~\ref{app:additional_benchmarks} to complement LLM-based evaluation results, we still believe the progress in better evaluating LLMs can lead to a more reliable demonstration of the effectiveness of our method.

Finally, as shown in Section~\ref{sec:ablation}, although \method is robust to moderate distribution mismatch, its performance still depends on how well the metadata captures the underlying instruction distribution. In practice, our collected seed instruction might differ from the actual test instructions. Or in the case that we directly create metadata from user specification, the users might change their mind at test time to send the model out-of-distribution instructions beyond the original metadata. As a consequence, \methodabbr may suffer performance degradation under distribution mismatch. As a remedy, we can constantly collect user instruction traffic or user feedback to update the generated data from \methodabbr, and continuously update the target LLM. 

We hope future work can leverage \methodabbr as a flexible data synthesis framework for LLM alignment, so that advances in the field can be integrated into \methodabbr to reduce its current limitations.

\appendix
\clearpage
\section{Appendix} \label{sec:appendix}
\subsection{Benchmark Details} \label{app:benchmark}
The details of the open-instruction following benchmarks are included below:
\begin{itemize}
    \item Evol-Instruct~\citep{xu2023wizardlm} includes 218 real-world human instructions from diverse sources such as online open-source projects, platforms, and forums. 
    \item Vicuna~\citep{vicuna2023} includes 80 diverse instructions generated by GPT-4 through prompt engineering.
    \item Self-Instruct~\citep{wang2022self} includes 252 expert-written instructions motivated by user-oriented applications.
    \item Koala~\citep{geng2023koala} includes 180 conversation-style real user instructions that were posted online.
\end{itemize}
All these benchmarks consist of English instructions from multiple categories or tasks. However, though sharing some common use cases such as general knowledge QA and coding, the coverage of the instructions in different benchmarks are indeed different. For example, ~\citet{xu2023wizardlm} discuss in detail how Evol-Instruct is different from Vicuna in instruction distribution. The difference between instruction distributions effectively mimic the practical scenario where we have different downstream tasks.

The details of the additional standard NLP benchmarks are included below:
\begin{itemize}
    \item MMLU~\citep{hendrycks2020measuring}, Massive Multitask Language Understanding, is a benchmark designed to measure capability of language models. It covers 57 subjects across STEM, the humanities, the social sciences, and more areas. We only use the test split for reporting the test results, and report the average score across all tasks.
    \item BBH~\citep{suzgun2022challenging}, BIG-Bench-Hard, includes 23 challenging BIG-Bench tasks that prior language models did not outperform average human-raters.
\end{itemize}

All benchmarks are publicly available for non-commercial research purposes, and we strictly limit their usage in this research work. We also carefully check these datasets and make sure that no personal information is involved.

\subsection{Baseline Details} \label{app:baseline_details}
\noindent \textbf{Self-Instruct} \citep{wang2022self} generates instructions by prompting LLM with existing seed instructions as few-shot demonstrations. Here we randomly subsample the Alpaca~\citep{alpaca} dataset as seed instructions. Since Alpaca itself is based on Self-Instruct, using its subset as seed is a natural continuation of the Self-Instruct method.

\noindent \textbf{Alpagasus} \citep{chen2023alpagasus} selectively filters data using ChatGPT-based response quality evaluator. Closely following the original approach, we adopt the strategy upon instruction-response pairs generated by Self-Instruct.

\noindent \textbf{Tree-Instruct} \citep{zhao2023preliminary} improves instruction quality by prompting the LLM to implicitly complicate instruction through its semantic tree. Following the original paper, we use the subsampled Alpaca dataset as seed data. We set the number of tree nodes to 10 for best possible performance.

\noindent \textbf{WizardLM} \citep{xu2023wizardlm} iteratively complicates instructions by prompting the LLM with a set of pre-defined evolution operations. Given the popularity and effectiveness of WizardLM, we experiment it with two variants: the original version using Alpaca as seed data, and the enhanced version uses the same set of basic instructions generated from \method as seed data. We name the later variant as \textbf{WizardLM+} as its enhanced by components of our framework.

\subsection{Additional Implementation Details} \label{app:implementation_details}
We augment the metadata to 200 by mix-and-matching use cases and skills from different instructions. We randomly sample one use case from $\{u_i\}_{i=1}^{n}$, and pair it with one or more skills sampled without replacement from $\bigcup_{i=1}^{n}\vs_i$. Although most skills are generalizable between use cases, we still conduct manual sanity check to exclude unreasonable use case and skills pairs.
We align our hyperparameters for iteratively improving instructions via \sr with prior work~\citep{xu2023wizardlm}: We generate 4 rubrics and corresponding actions, and at each iteration, we randomly choose 1 action for improving instruction. For fair comparison with WizardLM, we also use at most 4 improve iterations for each instruction (we count basic prompt generation as the first iteration). For Contrastive Filtering, we always use the strong LLM itself as the scorer. We set the scoring scale to 10 and the filtering threshold to 3 for all experiments. We obtain the threshold by developing on the AlpacaEval~\citep{dubois2023alpacafarm} dataset. And we find this threshold works generally well across different settings. Moreover, for \llama-based models, using their Alpaca~\citep{alpaca} counterparts as the target LLM for response generation in \cf works better than the original model that is not instruction tuned. For metadata extraction, base instruction generation and Self-Rubrics, we use a inference temperature of 0.7. We set the maximum number of tokens for generation to 2048 for \llama-based models, and 1024 for \palm-based models due to API constraints. Moreover, although we set aside 20\% validation set for metadata extraction, we still report the performance on the full test set in the main paper, the reasons are as follows: (1) We observe removing the validation set from the full test benchmark will not change the relative superior performance of our method, the performance gap between our method and baselines remains almost the same. Therefore, we keep them in for better reproducibility. (2) By carefully checking the generated instructions, we notice that none of the generated instructions overlap with the original validation instructions, so no data leaking happens during the data generation process.

We conduct manual inspection on the generated data to make sure no personal information or offensive contents are generated.

\subsection{Training Details} \label{app:training_details}

For \llama-based models, we follow the practices in instruction tuning in prior works~\citep{zhou2023lima,chen2023alpagasus}. We use AdamW optimizer with $\beta_1 = 0.9, \beta_2 = 0.95$ to finetune the target model for 15 epochs, as suggested by \citet{zhou2023lima} for smaller data size. 
We set the initial learning rate to $1\times10^{-5}$ and linearly decaying to $1\times10^{-6}$ by the end of training. We set per GPU batch size to 8, which is equivalent to a total batch size of 64,
as we use 8 A100 GPUs for training. The maximum token length is set to 2048.

For \palm-based models, we follow the default instruction tuning setting on Google Cloud's LLM tuning web UI. We set the number of tuning steps to 2000, the learning rate multiplier to 1, and use the TPU training option.

\begin{table}[t]
\small
    \centering
    \caption{Additional results on standard benchmarks.}
    \vspace{-3mm}
    \scalebox{0.98}{
        \begin{tabular}{c||ccc}
        \toprule
        \bf Methods &  \bf BBH & \bf MMLU & \bf Average \\
        \midrule
        \llama-7B & 30.93 & 35.17 & 33.05 \\
        Alpagasus & 31.55 & 36.46 & 34.01 \\
        WizardLM+ & 31.72 & 37.89 & 34.81 \\
        \methodabbr (ours) & \bf 32.60 & \bf 42.67 & \bf 37.64 \\
        \bottomrule
        \end{tabular}
    }
    \label{tab:standard_benchmark}
\end{table}

\subsection{Detailed Comparison Results} \label{app:detailed_comparison}
We show the details of pairwise comparison on Evol-Instruct benchmark with \llama-based models, as a demonstration of how CRR faithfully reflects the capability of the target LLMs trained by different methods. In Table~\ref{tab:results-details}, we observe that number of ties dominates the results and the number of wins are scarce. We attribute it to the fact that the target model is essentially distilling knowledge from the strong model. As a result, most of the time, the instruction-tuned target model is only able to respond as good as the strong model, through the lens of the LLM-based evaluator.

\begin{table*}[htbp]
\small
\centering
\caption{Detailed comparison results with \llama-based models on Evol-Instruct benchmark. Each method trains a target model based on \llama-7B or -13B, and compares against the strong model, Gemini-Pro. Capacity Recovery Ratio (\%), $\texttt{CRR} = \frac{\texttt{wins} + \texttt{ties}}{\texttt{total comparisons}}$.
}
\scalebox{0.92}{
\begin{tabular}{l||cccc||cccc}
\toprule
\multirow{2}[4]{*}{\textbf{Methods}} & \multicolumn{4}{c||}{\textbf{\llama-7B vs. Gemini-Pro}} & \multicolumn{4}{c}{\textbf{\llama-13B vs. Gemini-Pro}} \\
\cmidrule{2-9}      & \textbf{Wins} & \textbf{Ties} & \textbf{Losses} & \textbf{CRR} & \textbf{Wins} & \textbf{Ties} & \textbf{Losses} & \textbf{CRR} \\
\midrule
Self-Instruct & 17 & 140 & 61 & 72.02 & 29 & 136 & 53 & 75.69 \\
Alpagasus & 17 & 147 & 54 & 75.23 & 26 & 148 & 44 & 79.82 \\
Tree-Instruct & 23 & 141 & 54 & 75.23 & 26 & 154 & 38 & 82.57 \\
WizardLM & 19 & 143 & 56 & 74.31 & 30  & 149 & 39 & 82.11 \\
WizardLM+ & 19  & 146  & 53  & 75.69 & 31 & 153 & 34 & 84.40 \\
\methodabbr (ours) & \bf 29 & \bf 145 & \bf 44 & \bf 79.82  &  \bf 35 & \bf 154 & \bf 29 & \bf 86.70 \\
\bottomrule
\end{tabular}%
}
\label{tab:results-details}%
\end{table*}%

\subsection{Consistency between LLM-based Evaluators} \label{app:gpt4}
\begin{table*}[h]
\small
\centering
\caption{Performance gap to Self-Instruct in terms of CRR on Evol-Instruct, evaluated by ChatGPT and GPT4, respectively. Each method trains a target model based on \llama-7B or -13B, and compares against the strong model, Gemini-Pro. We observe two LLM-based automatic evaluators yields consistent results.
}
\scalebox{0.92}{
\begin{tabular}{l||cc||cc}
\toprule
\multirow{2}[4]{*}{\textbf{Methods}} & \multicolumn{2}{c||}{\textbf{\llama-7B vs. Gemini-Pro}} & \multicolumn{2}{c}{\textbf{\llama-13B vs. Gemini-Pro}} \\
\cmidrule{2-5} & \textbf{ChatGPT} & \textbf{GPT4} & \textbf{ChatGPT} & \textbf{GPT4}  \\
\midrule
Self-Instruct & 0.00 & 0.00 & 0.00 & 0.00  \\
Alpagasus & +3.21 & +1.38  & +4.13 & +1.83  \\
Tree-Instruct & +3.21 & +2.29 & +6.88 & +4.59 \\
WizardLM & +2.29 & +0.46 & +6.42 & +3.21 \\
WizardLM+ & +3.67 & +2.29 & +8.72 & +5.50 \\
\methodabbr (ours) & \bf +7.80 & \bf +8.26 & \bf +11.01 & \bf+8.72  \\
\bottomrule
\end{tabular}%
}
\label{tab:chatgpt-gpt4}%
\end{table*}%

In the main paper, we use ChatGPT as the LLM judge for final evaluation, for its efficiency, price and accessibility for the community to reproduce our results. As pointed out in~\citep{vicuna2023}, LLMs evaluators, although largely consistent with human preferences, may have their own biases. Therefore, to make sure our experimental results are solid, we also use GPT-4 as the judge and compare against the performance gap in CRR between different baselines and the Self-Instruct method. The comparison results in Table~\ref{tab:chatgpt-gpt4} demonstrates the agreement of two LLM-based judges and confirms the superior performance of \methodabbr against comparing methods.

\subsection{Additional Benchmark Results} \label{app:additional_benchmarks}
To complement the performance result using LLM-based automatic evaluator, we also evaluate LLMs tuned with the top methods presented in Section~\ref{sec:main_results} on standard NLP benchmarks, MMLU~\citep{hendrycks2020measuring} and BBH~\citep{suzgun2022challenging}. We follow the same settings introduced in~\citep{wang2023far} without demonstrations or CoT~\citep{wei2022chain} prompt for evaluating the target models based on \llama-7B. For our method, we follow the same setting as in Evol-Instruction benchmark evaluation. We present the evaluation results in Table~\ref{tab:standard_benchmark} and use the performance of vanilla \llama-7B as a reference. We observe the same performance ranking of all methods as that in Table~\ref{tab:results-llama} where we use LLM-based automatic evaluator. The consistency between two different evaluation approaches indicates the reliability of LLM-based evaluator in terms of demonstrating relative performance of competing methods.

\begin{figure*}[!t]
    \centering
    \includegraphics[width=0.98\linewidth]{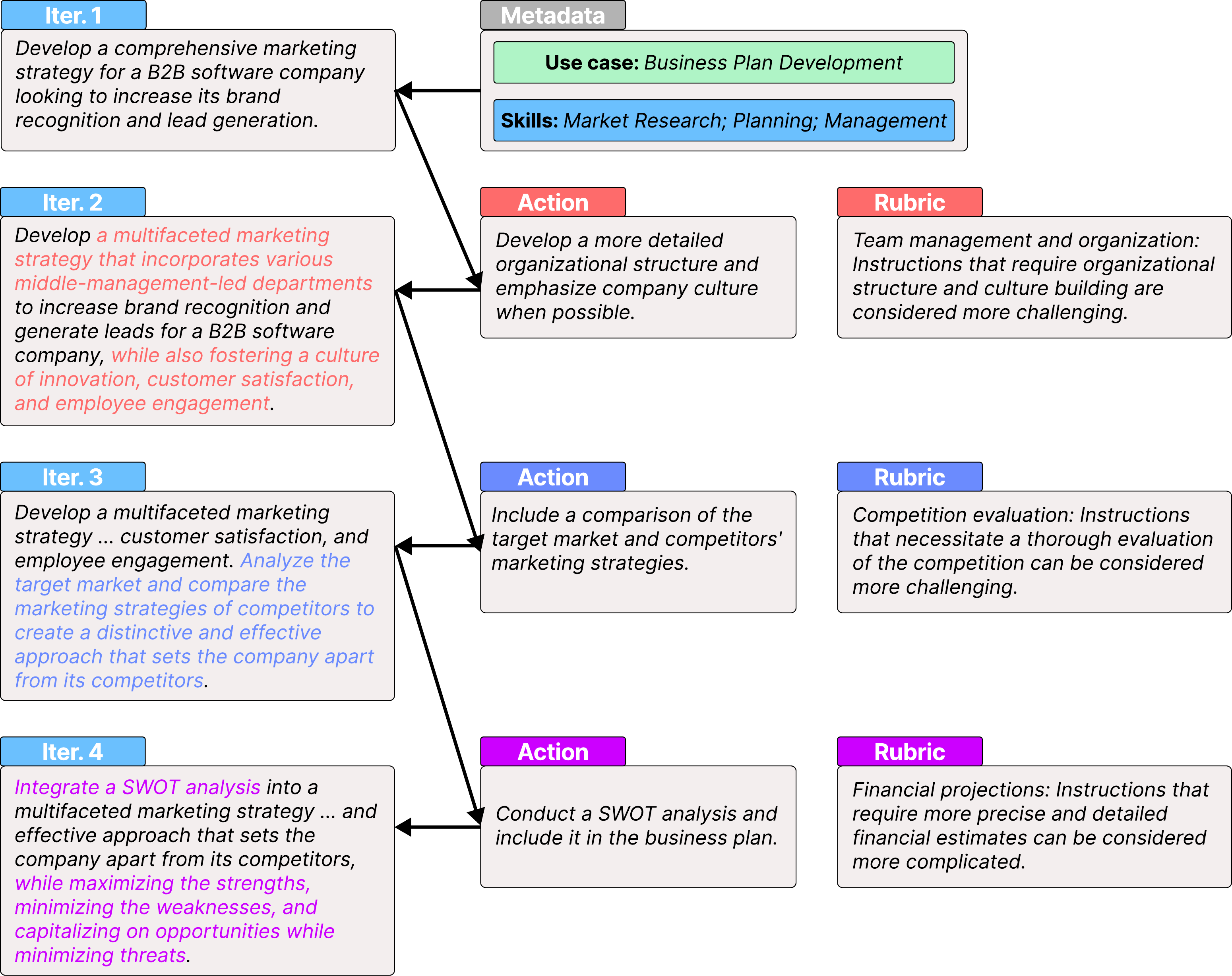}
    \caption{Case study on the instruction improvement process of \methodabbr. Repetitive instructions are omitted to save space.}
    \label{fig:case_study}
\end{figure*}

\subsection{Case Study} \label{app:case_study}
We present a case study in Figure~\ref{fig:case_study} to show an iterative tailoring process from instruction metadata to the final high-quality prompt. In practice, the iteration may terminate earlier by the \cf process. We observe that \sr is able to tailor rubrics and actions according to the given metadata. Interestingly, the actions generated by LLM seems very domain-specific. For example, the \emph{SWOT analysis} in the last action may even be hard for non-expert human annotators to come up with. Moreover, the colored texts in instructions demonstrate that LLM is able to follow the actions quite precisely to refine the instructions.

\subsection{Prompt Templates for \methodabbr} \label{app:prompts}
We present all prompt templates here in the appendix for better reproducibility. In particular, we list the correspondence between prompt templates and their usages as follows for quick reference:
\begin{itemize}
    \item Figure~\ref{fig:metadata}: Encoding instructions into metadata, including use case and transferable skills.
    \item Figure~\ref{fig:base_generation}: Decoding instruction metadata into basic instructions that are relatively simple in structure.
    \item Figure~\ref{fig:rubric_action_generation}: Generating rubrics to judge how challenging an instruction is, and actions to improve the instruction based on the given metadata.
    \item Figure~\ref{fig:instruction_improvement}: Improving the input instruction by following one of the generated actions.
    \item Figure~\ref{fig:cf_scorer}: Comparing the responses quality from the target and strong LLMs. Adapted from the Vicuna-style pairwise comparison prompt by removing the explanation part.
    \item Figure~\ref{fig:vicuna_judge}: Automatic evaluation using LLM (\eg, ChatGPT, GPT-4) as the judge. Following the templates in~\citep{vicuna2023,chen2023alpagasus}
\end{itemize}
All prompts are zero-shot except for the first encoding prompt in Figure~\ref{fig:metadata}, which utilizes few-shot demonstrations to showcase the LLM a rough granularity of the task and skills. Also, we choose these prompts as they work quite well in practice. And we believe recent prompt optimization techniques~\citep{fernando2023promptbreeder,yang2023large} can be incorporated seamlessly into our framework, and we leave them as future work.

\begin{figure*}[h] 
\begin{minted}[frame=single,
               framesep=2mm,
               fontsize=\small,
               tabsize=4]{text} 
I want you to act as an instruction analyzer. 
Given an instruction, you should recognize its use case and the skills (or knowledge) 
required for a large language model (LLM) to answer the question. 
Generate the use case and skills required without any explanation. 
List at most 3 skills, each skill should be transferable, so that LLM can leverage them to answer 
similar questions. 
Avoid using "skill", "knowledge" to describe a skill, and each skill should be concise (2-3 words).
Follow the examples below to analyze the given instruction.

#Example 1#
As a sports commentator, describe the winning play in the final seconds of a championship game.
Use case: creative writing
Skills: role-play, sports

#Example 2#
How to read a large file (> 2T) using python?
Task: code generation
Skills: python

#Example 3#
The method section of your paper is too brief and does not explain how your proposed model works 
in detail. How can you provide more details of the hierarchical encoder and the cascaded selectors, 
such as their architectures, inputs, outputs, and parameters?
Task: general knowledge question answering
Skills: academic writing, machine learning

<input instruction>
<output metadata>
\end{minted}
\caption{Prompt template to encode the input into metadata, consisting of its use case and transferable skills.} \label{fig:metadata}
\end{figure*}

\begin{figure*}[h] 
\begin{minted}[frame=single,
               framesep=2mm,
               fontsize=\small,
               tabsize=4]{text} 
I want you to act as an instruction writer.
Your objective is to write <number of instructions> instructions that must be reasonable 
and must be understood and responded by humans. 
The generated instructions should be diverse enough while following the constraints below:

Use case of the instructions: <use case>
Skills required to respond to the instructions: <skills>

Generate the instructions without answering in numbered bulletin points.

<output instructions>
\end{minted}
\caption{Prompt template to generate instructions from metadata.} \label{fig:base_generation}
\end{figure*}

\begin{figure*}[h] 
\begin{minted}[frame=single,
               framesep=2mm,
               fontsize=\small,
               tabsize=4]{text} 
I want you to act as a instruction judge with domain expertise. 
Your job is to generate <number_of_rubrics> domain specific rubrics to assess the difficulty and 
complexity based on the use case of the instruction, and skills required to respond to it.
The generated rubrics should be clear, concise and unambiguous.
Based on the generated rubrics, generate corresponding actions to improve an instruction by
making it more challenging.

The use case of the instruction: <use case>.
The skills required to solve the instruction: <skills>.

Generate the domain-specific rubrics and actions without explanation in numbered bulletin points:

<output rubrics>
<output actions>
\end{minted}
\caption{Prompt template to generate actions to improve instructions based on instruction metadata.} \label{fig:rubric_action_generation}
\end{figure*}

\begin{figure*}[h] 
\begin{minted}[frame=single,
               framesep=2mm,
               fontsize=\small,
               tabsize=4]{text} 
I want you to act as a instruction improver with domain expertise. 
Your job is to make the given instruction more challenging following the given improving action 
item, and the generated instruction should be reasonable and self-consistent. 
Do not directly copy words or phrases in the action.

Improving action: <action>
Input instruction: <input instruction>

Improved instruction: <output instruction>
\end{minted}
\caption{Prompt template to improve instructions following generated actions.} \label{fig:instruction_improvement}
\end{figure*}

\begin{figure*}[h] 
\begin{minted}[frame=single,
               framesep=2mm,
               fontsize=\small,
               tabsize=4]{text}
You are a helpful and precise assistant for checking the quality of the answer.

<Question>
[The Start of Assistant 1's Answer]
<answer_1>
[The End of Assistant 1's Answer]
[The Start of Assistant 2's Answer]
<answer_2>
[The End of Assistant 2's Answer]

We would like to request your feedback on the performance of two AI assistants in response to 
the user question displayed above. 
Please rate the helpfulness, relevance, accuracy, level of details of their responses. Each 
assistant receives an overall score on a scale of 1 to 10, where a higher score indicates 
better overall performance.
Please only output a single line containing only two values indicating the scores for Assistant 1 
and 2, respectively. The two scores are separated by a space.
Please avoiding any potential bias and ensuring that the order in which the responses were 
presented does not affect your judgment.
\end{minted}
\caption{Prompt template used in \cf to compare the responses of the strong and the target LLMs. We directly use the strong LLM with this template as the scorer $S$ to avoid additional costs from calling a third-party LLM.} \label{fig:cf_scorer}
\end{figure*}

\begin{figure*}[h] 
\begin{minted}[frame=single,
               framesep=2mm,
               fontsize=\small,
               tabsize=4]{text}
System: You are a helpful and precise assistant for checking the quality of the answer.

User:
<Question>
[The Start of Assistant 1's Answer]
<answer_1>
[The End of Assistant 1's Answer]
[The Start of Assistant 2's Answer]
<answer_2>
[The End of Assistant 2's Answer]

We would like to request your feedback on the performance of two AI assistants in response to 
the user question displayed above. 
Please rate the helpfulness, relevance, accuracy, level of details of their responses. Each 
assistant receives an overall score on a scale of 1 to 10, where a higher score indicates 
better overall performance.
Please first output a single line containing only two values indicating the scores for Assistant 1 
and 2, respectively. 
The two scores are separated by a space. In the subsequent line, please provide a comprehensive
explanation of your evaluation, avoiding any potential bias and ensuring that the order in which 
the responses were presented does not affect your judgment.
\end{minted}
\caption{Prompt template for automatic evaluation using LLM (\eg, ChatGPT, GPT-4) as the judge.} \label{fig:vicuna_judge}
\end{figure*}

\end{document}